\pdfoutput=1

\documentclass[11pt]{article}

\usepackage[preprint]{acl}

\usepackage{times}
\usepackage{latexsym}

\usepackage[T1]{fontenc}

\usepackage[utf8]{inputenc}

\usepackage{microtype}

\usepackage{inconsolata}

\usepackage{graphicx}

\usepackage{amsmath}
\usepackage{amssymb}
\usepackage{booktabs}
\usepackage{subcaption}
\usepackage{diagbox}

\usepackage{pbox}
\usepackage{makecell}
\usepackage{multirow}
\usepackage{graphicx}

\usepackage[inline]{enumitem}

\title{MoreHopQA: More Than Multi-hop Reasoning}

\author{
Julian Schnitzler,$^{\thefootnote{*} \;1}$
Xanh Ho,$^{\thefootnote{*} \; 2, 3}$
Jiahao Huang,$^{\thefootnote{*} \; 4}$ \\
\textbf{Florian Boudin},$^{3,5}$
\textbf{Saku Sugawara},$^3$\and
\textbf{Akiko Aizawa}$^{2,3,4}$ \\
$^1$EPFL, Lausanne, Switzerland\\
$^2$The Graduate University for Advanced Studies, Kanagawa, Japan\\
$^3$National Institute of Informatics, Tokyo, Japan \\
$^4$The University of Tokyo, Japan \hspace{1cm} $^5$JFLI, CNRS, Nantes Université, France \\
{\tt julian.schnitzler@epfl.ch} \hspace{1cm}
{\tt \{xanh, saku, aizawa\}@nii.ac.jp}  \\
{\tt jiahao-huang@g.ecc.u-tokyo.ac.jp} \hspace{1cm}
{\tt florian.boudin@univ-nantes.fr}
}

\begin{document}
\maketitle

\begingroup\def\thefootnote{*}\footnotetext{Equal contribution.}\endgroup

\begin{abstract}
Most existing multi-hop datasets are extractive answer datasets, where the answers to the questions can be extracted directly from the provided context. 
This often leads models to use heuristics or shortcuts instead of performing true multi-hop reasoning.
In this paper, we propose a new multi-hop dataset, \texttt{MoreHopQA}, which shifts from extractive to generative answers.
Our dataset is created by utilizing three existing multi-hop datasets: HotpotQA, 2WikiMultihopQA, and MuSiQue. 
Instead of relying solely on factual reasoning, we enhance the existing multi-hop questions by adding another layer of questioning that involves one, two, or all three of the following types of reasoning: commonsense, arithmetic, and symbolic.
Our dataset is created through a semi-automated process, resulting in a dataset with 1,118 samples that have undergone human verification. 
We then use our dataset to evaluate five different large language models: Mistral 7B, Gemma 7B, Llama 3 (8B and 70B), and GPT-4.
We also design various cases to analyze the reasoning steps in the question-answering process. 
Our results show that models perform well on initial multi-hop questions but struggle with our extended questions, indicating that our dataset is more challenging than previous ones.
Our analysis of question decomposition reveals that although models can correctly answer questions, only a portion—38.7\% for GPT-4 and 33.4\% for Llama3-70B—achieve perfect reasoning, where all corresponding sub-questions are answered correctly.\footnote{Our data and code are available at \url{https://github.com/Alab-NII/morehopqa}}
\end{abstract}

\section{Introduction}

\begin{figure}[tb!]
\centering
    \includegraphics[width=\linewidth]{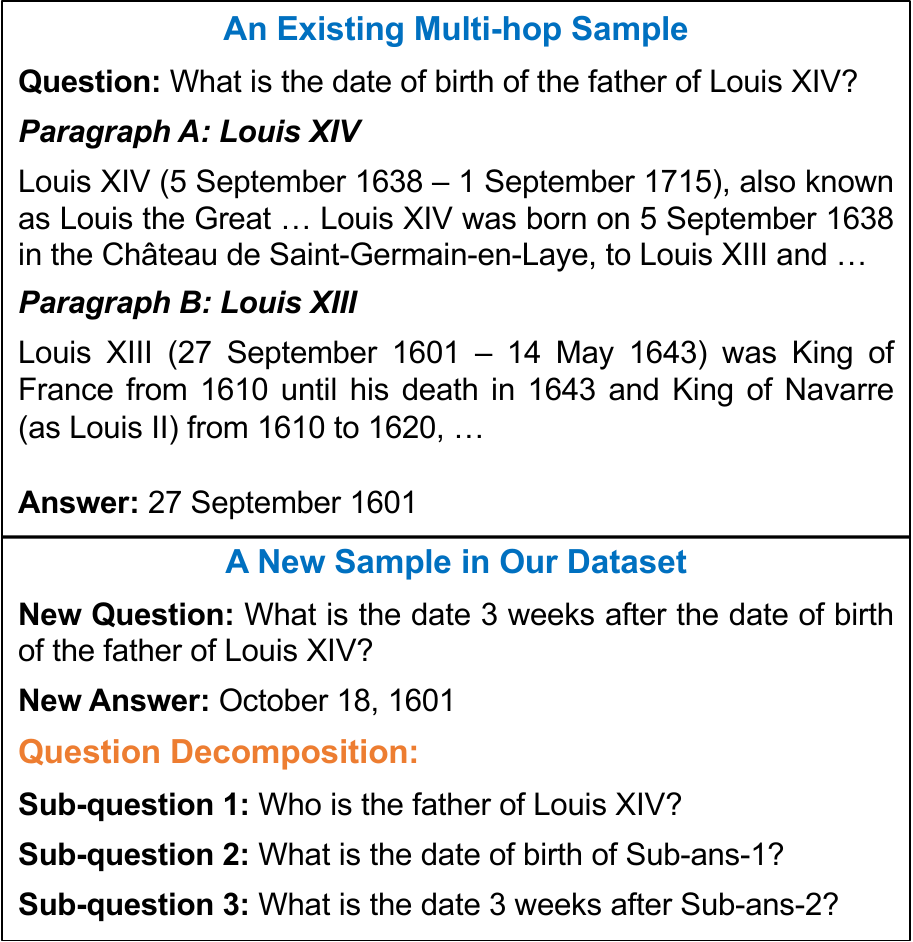}
    \caption{An example of our dataset.
    Our new question is created by extending the initial 2-hop question, which ensures that the new answer is generative.
    }
    \label{fig_example}
\end{figure}

Multi-hop Question Answering (QA) requires a model to retrieve, extract, and connect pieces of evidence from multiple paragraphs to answer a question~\cite{welbl-etal-2018-constructing,yang-etal-2018-hotpotqa}.
By harnessing the reasoning abilities of models, this task provides valuable insights into evaluating their capabilities in understanding natural language and tackling complex tasks.
For this reason, multi-hop QA has received much attention over the past few years, prompting the creation of several benchmark datasets such as HotpotQA~\cite{yang-etal-2018-hotpotqa}, 2WikiMultihopQA~\cite[2Wiki;][]{ho-etal-2020-constructing}, MuSiQue~\cite{trivedi-etal-2022-musique-custom}, MQuAKE~\cite{zhong-etal-2023-mquake}, MRKE~\cite{wu2024mrke}, or FanOutQA~\cite{zhu2024fanoutqa}.

While existing multi-hop QA datasets have been instrumental in evaluating the reasoning capabilities of Large Language Models (LLMs), they suffer from several limitations.
The first limitation concerns the type of answers found in these datasets.
Indeed, most of the answers are extractive, meaning they can be directly extracted from the supporting paragraphs provided as context.
Such answers may incentivize models to generate answers through heuristics or reasoning shortcuts~\cite{min-etal-2019-compositional,Geirhos2020a,ho2023survey}, rather than engaging in the expected multi-step reasoning task.
For example, questions asking about dates with supporting paragraphs containing only one possible date entity are likely to be guessed correctly by models.
The second limitation lies in the restricted range of reasoning types found in existing multi-hop datasets, which primarily focus on reasoning tasks involving common knowledge from Wikipedia.
Consequently, they neglect other forms of reasoning, such as arithmetic or symbolic reasoning, which are also crucial to consider when evaluating the reasoning capabilities of models~\cite{qiao-etal-2023-reasoning}.

In this paper, we aim to address these limitations by introducing MoreHopQA, a new dataset made of multi-hop questions whose answers cannot be simply extracted and instead require combining multiple types of reasoning.
Our approach involves extending questions from existing datasets with additional hops, thereby transforming their original answers into generative answers, which prevents them from being simply guessed by models (see Figure~\ref{fig_example}).
More specifically, our dataset features the following main aspects:
%
\begin{enumerate*}[label=\arabic*)]
    \item Answers are generative, requiring models to reason to derive the final answer.
    \item To answer questions in our dataset, models need to engage in multi-step reasoning first, followed by another type of reasoning (e.g.,~arithmetic).
    \item We provide explicit decompositions, that is, the set of sub-questions and sub-answers in the reasoning process from question to answer.
\end{enumerate*}
%
We argue that adopting generative answers and challenging models to perform additional types of reasoning beyond multi-hop questions can make the dataset more demanding for the models.

Our dataset creation process involves the following four steps:
\begin{enumerate*}[label=\arabic*)]
    \item \emph{Sample Selection} (\S\ref{subsec:sample-selection}), where we manually curated 2-hop samples from three existing multi-hop datasets (i.e.~HotpotQA, 2Wiki, and MuSiQue) according to three criteria: questions should be answerable, include sub-questions and sub-answers, and have properly formatted answers.
    \item \emph{Template Design} (\S\ref{subsec:template-design}), where we (the authors of this paper) collaboratively designed about 100 templates for creating new questions encompassing three types of reasoning (i.e.~arithmetic, commonsense, and symbolic) from five answer types (i.e.~person, place, organization, date and year).
    \item \emph{New Sample Generation} (\S\ref{subsec:new-sample-generation}), where we use our templates in conjunction with the selected 2-hop samples to automatically generate new samples.
    \item \emph{Human Verification} (\S\ref{subsec:human-verification}), where we ensure the quality of our new samples by asking a pool of annotators to label and revise them, resulting in a final dataset of 1,118 human verified samples.
\end{enumerate*}
We further validate the quality of our dataset by evaluating human performance on a subset of 150 samples, demonstrating that our new samples are both answerable and reasonable (\S\ref{sec_human_perform}).

\begin{figure*}[ht]
\centering
    \includegraphics[width=\textwidth]{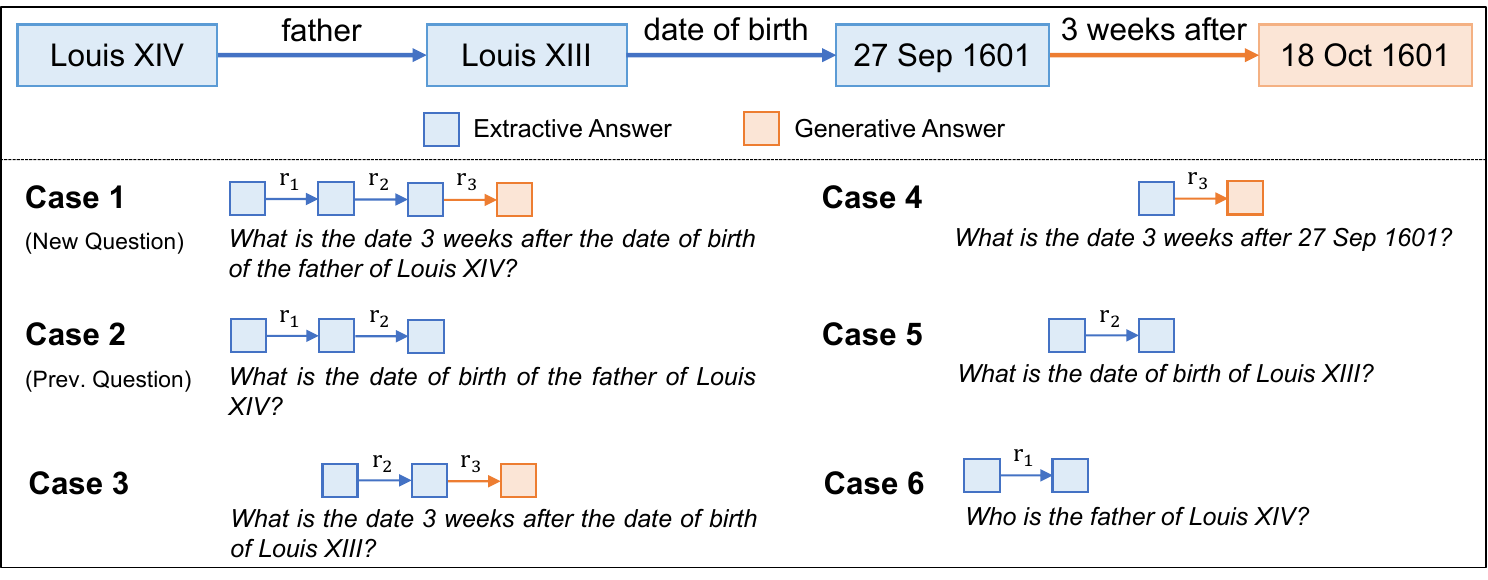}
    \caption{
    There are six cases in our analyses. 
    The first case is our newly generated question. The second case is the initial 2-hop question.
    We present the details of these cases in Appendix~\ref{app_data_analysis}.
    }
    \label{fig_overall}
\end{figure*}

We then use our dataset to evaluate the reasoning capabilities of five different LLMs: Mistral 7B, Gemma 7B, Llama 3 (8B and 70B), and GPT-4.
We conduct experiments using multiple prompting strategies, including zero-shot, few-shot, and Chain-of-Thought (CoT)~\cite{wei2022chainofthought}.
We leverage the explicit decompositions of the questions in our dataset to conduct an extensive error analysis (Figure~\ref{fig_overall}), 
precisely identifying where in the reasoning chain the models fail and highlighting which models resort to reasoning shortcuts.
Our results indicate that while the models perform well on the initial multi-hop questions, they struggle more with our extended questions. This suggests that our dataset presents a greater challenge compared to previous datasets.
Our analysis of question decomposition reveals that while models can correctly answer questions, only a small portion (38.7\% for GPT-4 and 33.4\% for Llama3-70B) achieve perfect reasoning, where all corresponding sub-questions are answered correctly.

In summary, our contributions are as follows: 

\begin{itemize}
    \item We create a more challenging dataset that shifts from extractive to generative, and, with the decompositions, allows for a better understanding of the reasoning capabilities of LLMs.
    
    \item We conduct extensive human verification and validation to ensure the quality of our dataset.

    \item We evaluate the performance of five LLMs and show that even state-of-the-art LLMs do not match human performance. We also find that while GPT-4 performs best, 
    only 38.7\% reach the state of perfect reasoning.
\end{itemize}

\section{Related Work}
\label{sec:related-work}

\subsection{Multi-hop QA Datasets}

The first multi-hop QA dataset, QAngaroo, was introduced by~\citet{welbl-etal-2018-constructing}. It consists of two sub-datasets, WikiHop and MedHop, and was constructed by leveraging both unstructured text sources (e.g.~Wikipedia or Medline) and structured data from external resources (e.g.~Wikidata or DrugBank). 
In the same year, \citet{talmor-berant-2018-web} introduced ComplexWebQuestions, a dataset derived from WebQuestionsSP~\cite{yih-etal-2016-value} that contains automatically generated questions revised by crowdworkers.
In the following years, HotpotQA~\cite{yang-etal-2018-hotpotqa}, $\mathrm{R^4C}$~\cite{inoue-etal-2020-r4c}, 2WikiMultihopQA~\cite{ho-etal-2020-constructing}, and MuSiQue~\cite{trivedi-etal-2022-musique-custom} were introduced, with a greater emphasis on explaining the QA process.
%
%
MQuAKe~\cite{zhong-etal-2023-mquake} and FanOutQA~\cite{zhu2024fanoutqa} are two recently proposed datasets. MQuAKe focuses on testing multi-hop reasoning for knowledge editing in LLMs, while FanOutQA focuses on creating complex listing questions.
However, many existing datasets only feature extractive answers and focus solely on multi-hop reasoning within Wikipedia text. 
In contrast, our dataset shifts from extractive to generative answers, requiring broader reasoning abilities for answering the questions.

\subsection{Multi-hop Analyses}
Due to the intricate nature of multi-hop questions, they are particularly useful for analyzing and evaluating the reasoning chains in the QA process.
\citet{tang-etal-2021-multi} utilized sub-questions in the QA process and conducted experiments on HotpotQA to determine whether multi-hop models could answer them successfully. They found that multi-hop models did not perform well on this task.

\citet{trivedi-etal-2020-multihop} used the connection between the two supporting facts to analyze the abilities of the models. 
They found that even with disconnections, the models could still answer the questions, revealing that the models can use heuristics or shortcuts to arrive at the answers.
In the shortcuts analyses, several previous works~\cite{min-etal-2019-compositional,chen-durrett-2019-understanding,jiang-bansal-2019-avoiding} also raised the issues about the multi-hop reasoning abilities of the models and the shortcuts in existing datasets.

Additionally, recent works~\cite{dua-etal-2022-successive,khot2023decomposed,press-etal-2023-measuring,zhou2023leasttomost} attempted to incorporate a question decomposition step into their prompts to improve model performance.
Prior to these studies, some works~\cite{talmor-berant-2018-web,min-etal-2019-multi,fu-etal-2021-decomposing-complex} showed that integrating question decomposition into their systems can lead to better performance and more explainable responses.
\citet{patel-etal-2022-question} showed that human decomposition improves performance on complex questions. 
However, \citet{wei-etal-2023-decompositions} showed that question decomposition does not help when there are more samples in the dataset.
%
Due to sparse benchmarks, drawing reliable conclusions about question decomposition is challenging.
Our dataset includes sub-questions and sub-answers, which could be valuable for future research on exploring the effectiveness of question decomposition.

\section{Dataset Creation Process}
\label{sec_dataset_creation}

\begin{figure*}[ht]
\centering
    \includegraphics[width=\textwidth]{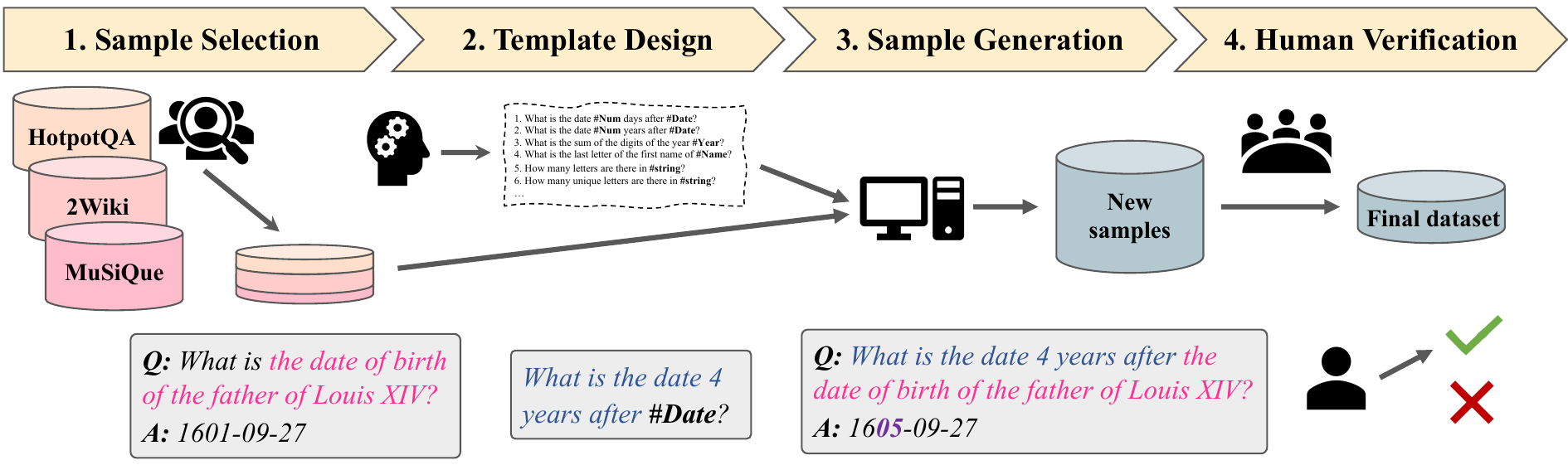}
    \caption{
   Our dataset creation process.
    }
    \label{fig_dataset_creation}
\end{figure*}

Our dataset creation process, illustrated in Figure~\ref{fig_dataset_creation}, consists of four main steps: 1)~sample selection, 2)~template design, 3)~new sample generation, and 4)~human verification.
%
We first describe each of these four steps and then provide detailed information about the final version of the dataset.

\subsection{Sample Selection}
\label{subsec:sample-selection}

Our new samples are derived from 2-hop questions found in three existing multi-hop datasets: HotpotQA~\cite{yang-etal-2018-hotpotqa}, 2Wiki~\cite{ho-etal-2020-constructing}, and MuSiQue~\cite{trivedi-etal-2022-musique-custom}.
%
To ensure the quality of our dataset, we defined three criteria for selecting the initial 2-hop samples:
%
%
%
%
1)~\textbf{Answerability}: all 2-hop questions should be answerable, that is, the answer must be found in the supporting paragraphs.
2)~\textbf{Decomposition}: initial 2-hop samples should contain a list of sub-questions and sub-answers.
3)~\textbf{Format}: we categorized the initial 2-hop samples based on their answer type, such as person name, date, year, or location, and applied specific requirements to each group. For example, dates should be fully formatted (comprising day, month, and year), while person names should include both the first and last names.
Herein, we describe the methodology we applied for selecting the initial samples from each dataset.

\paragraph{HotpotQA}
%
%
Since the original HotpotQA lacks sub-questions and sub-answers, we relied on \citet{tang-etal-2021-multi}, who annotated 1,000 samples with them.
From this pool, we manually curated a subset of samples, discarding those that are difficult to understand or have answers in an incorrect format, and annotated each sample with its corresponding answer type.
Notably, we observed that the format of the answers for the place type was inconsistent, making it difficult to integrate with templates, so we decided to exclude them.
%
We obtained 48, 47, and 19 samples with answer types of person, year, and date, respectively.


\paragraph{2Wiki}
We selected the bridge questions from the development set as our initial samples.
Based on the relation type of the second triple in the reasoning chain, we classified the samples into five answer types: place, person, year, date, and string.
Since questions in 2Wiki are automatically generated, we manually reviewed 400 samples to check their answerability and decide whether to use them.
For instance, we opted to exclude questions with answers of the string type, as they often have multiple valid answers.
We obtained 120, 114, 69, and 11 samples for place, date, person, and year, respectively.

\paragraph{MuSiQue}
We selected the samples from the development set with a structured format (similar to a triple format) for the second hop in the question decomposition process.
Based on the relation information of the second hop, we automatically annotated the answer types of the samples, resulting in 105, 99, and 22 samples for person, place, and organization, respectively. 
%
We observed that a substantial number of samples in MuSiQue have multiple answers, being either explicitly indicated in the dataset (\texttt{answer\_aliases} field) or identified during our manual verification process.
%
Because our new answers are based on the answers to the 2-hop questions, we do not include these samples in our dataset.
As a result, we obtain 17, 14, and 3 samples for person, place, and organization, respectively.
We present examples of these issues in Appendix \ref{appendix_data_sec_musi}, which further explains why the final number of samples drawn from MuSiQue is small.

\subsection{Template Design}
\label{subsec:template-design}

We, the authors of this paper, collaboratively designed 97 templates for creating the new questions in our dataset.
%
%
%
Multiple templates were designed for each answer type, with the purpose of creating new questions whose new answers are generative, meaning they can not be simply extracted from the supporting paragraphs.
%
%
For example, regarding the date answer type, we can ask about the next day, next month, next week, next year, or any other gap relative to the current date.
Another example for the person name answer type, we can ask about the first letter of the first name, the last letter of the first name, or the concatenation of the first letter and last letter of the first name.
As discussed in the Introduction, we conjecture that extractive answers are easy for models to identify, potentially leading to their tendency to rely on heuristics and shortcuts in the QA process.
Here, we purposely crafted our templates to address that issue, adding one extra hop to the initial 2-hop question to make the new answer a generative type.

In \citet{qiao-etal-2023-reasoning}, five types of reasoning are explored: arithmetic reasoning, commonsense reasoning, symbolic reasoning, logical reasoning, and multimodal reasoning.
We designed our templates to encompass the first three types of reasoning, but not extend to logical or multimodal reasoning due to the nature of the samples we use (multi-hop questions in the Wikipedia domain).
Our templates cover all three of these reasoning types individually, as well as various combinations thereof.
Some templates rely on a single type of reasoning, while others require two or three types.
%
%
Each template is labeled with its corresponding reasoning type(s), and we also indicate the number of hops required to answer the new questions.
If the number of required hops exceeds one, we include a list of sub-questions and their corresponding sub-answers.

\subsection{New Sample Generation}
\label{subsec:new-sample-generation}

We use the list of templates in conjunction with the selected 2-hop samples to generate new samples for our dataset.
This involves creating both a new question and a new answer for each pair of template and 2-hop sample.
To generate a new question, we combine our templates with the noun phrases extracted from the initial 2-hop questions.
For example, given the question [\,\textit{What is the date of birth of the father of Louis XIV?}\,] and our template [\,\textit{What is the date one week after \#Date?}\,], we first extract the noun phrase of the question [\,\textit{the date of birth of the father of Louis XIV}\,].
Next, we replace the special token \textit{\#Date} in our template by this noun phrase to get [\,\textit{What is the date one week after the date of birth of the father of Louis XIV?}\,].
We also incorporate another special token \textit{\#Num} for numerical quantities, allowing us to choose various values (e.g.,~one week, two weeks) when generating new questions.
The 2-hop questions in 2Wiki and MuSiQue are well-structured, allowing us to extract their noun phrases using rule-based methods.
However, as the HotpotQA questions are crowdsourced, we resort to manual annotation to accurately identify the noun phrase of each question.
%
To obtain the new answer, we use code to perform the operations on the initial 2-hop answer corresponding to the template (e.g.,~adding one week).
An example of a generated sample is provided in Appendix~\ref{appendix_data_sec_our_data}.

From 114, 314, and 34 samples in HotpotQA, 2Wiki, and MuSiQue, respectively, we generate 1,497, 2,617, and 373 new samples.
There are four answer types in our dataset: date, number, string, and letter.
Statistics about the number of samples for each type are presented in Table~\ref{tab_dataset}.
An example question for each answer type is provided in Appendix~\ref{app_data_analysis}.

\subsection{Human Verification}
\label{subsec:human-verification}

After completing the previous steps, we have generated a total of 4,487 new samples.
Our focus now shifts to ensuring the quality of our dataset, as these newly generated questions may exhibit issues stemming from our template-based approach.
We extracted a subset of 1,408 randomly selected new samples for human verification and tasked 10 annotators (students and researchers in NLP, including the authors) with verifying and, if necessary, modifying the generated questions.
%
%
The human verification process involves labeling the new questions with one of the following three labels:
[\textbf{OK}] the question is acceptable and requires no changes;
[\textbf{Modified}] the question had flaws that were corrected through modifications;
and [\textbf{Issue}] the question has significant problems that remain despite attempts to modify it.
The guidelines and the annotation interface are provided in Appendix~\ref{appendix_data_sec_verification}.
%
%
Out of the 1,408 samples that were verified, 919 were labeled as OK, 408 as Modified, and 81 as Issue.
Questions labeled as Issues were double-checked, and those deemed unusable (e.g.,~initial 2-hop question having multiple answers) were discarded from our final dataset.

\subsection{Final Dataset}

After the human verification process, we are left with 1,118 new samples.
Statistics for the number of samples for each answer type in our final dataset are presented in Table~\ref{tab_dataset}.
In addition to the subset that underwent human verification, we also release the remaining subset of 2,502 samples without human verification.
%
For this latter subset, we automatically filtered out the samples derived from questions marked as erroneous through the human verification process, aiming to enhance its overall quality. 
Our dataset information is in English.

\begin{table}[tb!]
\resizebox{\columnwidth}{!}{%
    \centering
    \begin{tabular}{l r r r r r}
\toprule
    \textbf{Dataset} & \textbf{Date} & \textbf{Number} & \textbf{String} & \textbf{Letter} & \textbf{Total} \\
\midrule
    HotpotQA  & 76  & 1,070    & 304   & 47     & 1,497         \\
    2Wiki   & 567    & 1,453   & 528     & 69     & 2,617    \\
    MuSiQue   & 17  & 225  & 114  & 17   & 373  \\ 
\midrule 
    MoreHopQA {\small w/ hv}  &   216   &   663   &   196    &  43  &  1,118  \\
    MoreHopQA {\small w/o hv} &   436   &  1,526    &   479    & 61   &  2,502  \\
\bottomrule
\end{tabular}

}
    \caption{
    Statistics showing the number of generated samples for each answer type in our dataset.
    MoreHopQA {\small w/ hv} indicates the version with human verification.
    %
    }
    \label{tab_dataset}
\end{table}

\section{Dataset Quality Assessment}
\label{sec_human_perform}
To further validate the quality of our dataset and provide an estimate of human performance, we tasked the same pool of annotators as in \S\ref{subsec:human-verification} with answering a randomly selected subset of 150 samples.
Each sample consists of a question and two supporting (gold) paragraphs.
The task of the annotators is to answer the given questions.
Each sample is annotated by two separate annotators.
Since our aim is to assess the reasoning abilities of the process rather than focusing on its retrieval components, we do not include distractor paragraphs.

We calculate three distinct metrics: the average human performance, the human upper bound, and the inter-annotator agreement.
Following~\cite{yang-etal-2018-hotpotqa,ho-etal-2020-constructing}, the upper bound is computed as the average of maximum exact match (EM) for each sample.
We obtain scores of 84.3, 94.0, and 76.7 for these three metrics, respectively. 
The notably high human performance scores, encompassing both the average and upper bound, serve as strong indicators of the quality of the dataset. 
%
Notably, the human performance average score sets a benchmark for the expected model performance.
Furthermore, the inter-annotator agreement score, although slightly lower, remains within an acceptable range, affirming the consistency and reliability of our dataset.

\section{Experiments}
\label{sec_exp}

\begin{figure*}[ht!]
    \centering
    \includegraphics[width=0.99\textwidth]{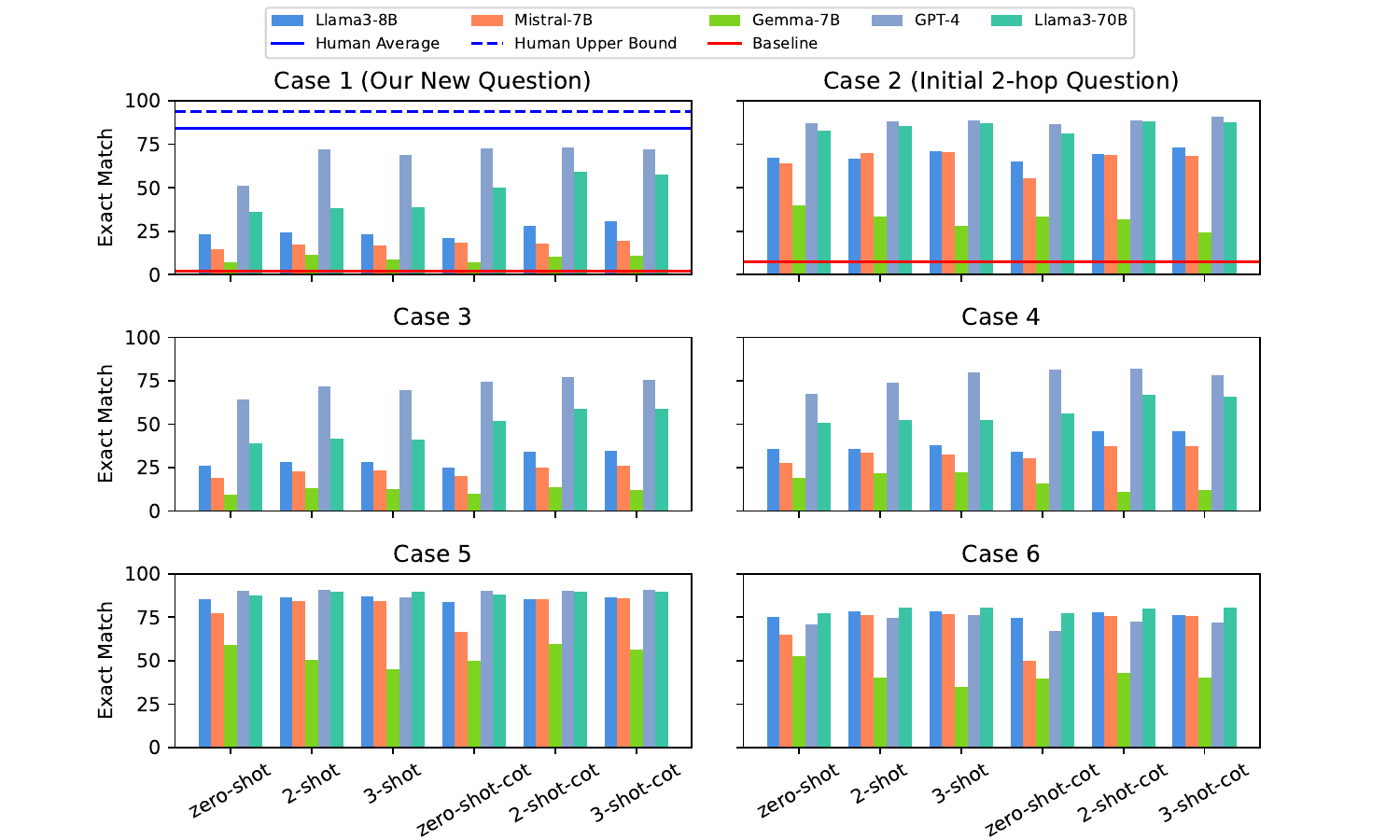}
    \caption{Performance (EM scores) of the models on our dataset. 
    }
    \label{fig:results}
\end{figure*}

\subsection{Experimental Settings}

\paragraph{Models}

We compare the performance of several instruction-fine-tuned auto-regressive LLMs on our dataset.
To represent a variety of current models in terms of size and fine-tuning, we chose Llama-3-8B-Instruct and Llama-3-70B-Instruct from the Llama-3 family of models~\cite{llama3modelcard}, as well as Mistral-7B-Instruct-v0.3~\cite{jiang2023mistral}, Gemma-7B~\cite{gemmateam2024gemma}, and GPT-4 Turbo \cite{openai2024gpt4}.

\paragraph{Prompting}

Following the results from \citet{kojima2022zeroshot} and \citet{wei2022chainofthought}, we compare the performance using zero-shot and few-shot prompting with 2 and 3 shots, as well as CoT prompting with zero, 2, and 3 shots. 
For comparability, we use the same user prompts for all models. The only variation in our prompting setup is the inclusion of a system prompt, which is applied when specified by the model's authors in its Hugging Face model card. We select the few-shot examples from our dataset in such a way that the answer types of the examples match those of our question, while ensuring that none of the answers to the subquestions are revealed in the prompt.

\paragraph{Baseline}
Following previous work on detecting potential reasoning shortcuts in datasets~\cite{sugawara-etal-2018-makes,trivedi-etal-2022-musique-custom}, we run an artifact-based baseline with Llama-8B.
In this baseline, we only use the two words from the question (e.g., ``when was'' or ``how many'').

\paragraph{Evaluation}

We follow the general approach of evaluating multi-hop QA tasks as presented in~\cite{yang-etal-2018-hotpotqa}, and additionally run postprocessing on the generated model output to extract the final answer, depending on the expected type of the answer.
When prompting, we ask the model to give the final answer between two <answer> tags, and parse the string between those as the model's final answer. We then attempt to convert this string into the respective built-in python datatype for the answer type, either directly or with the help of Named Entity Recognition, and convert it back to a default string representation. We then report the EM and F1 scores on the tokens between the preprocessed ground-truth answer and the postprocessed model-generated answer.

\subsection{Results}

The performance (EM scores) of all models on our dataset are presented in Figure \ref{fig:results}. 
We present both EM and F1 scores in Appendix~\ref{app_experiment_result}.

\begin{figure*}[!ht]
    \centering
    \includegraphics[width=0.91\textwidth]{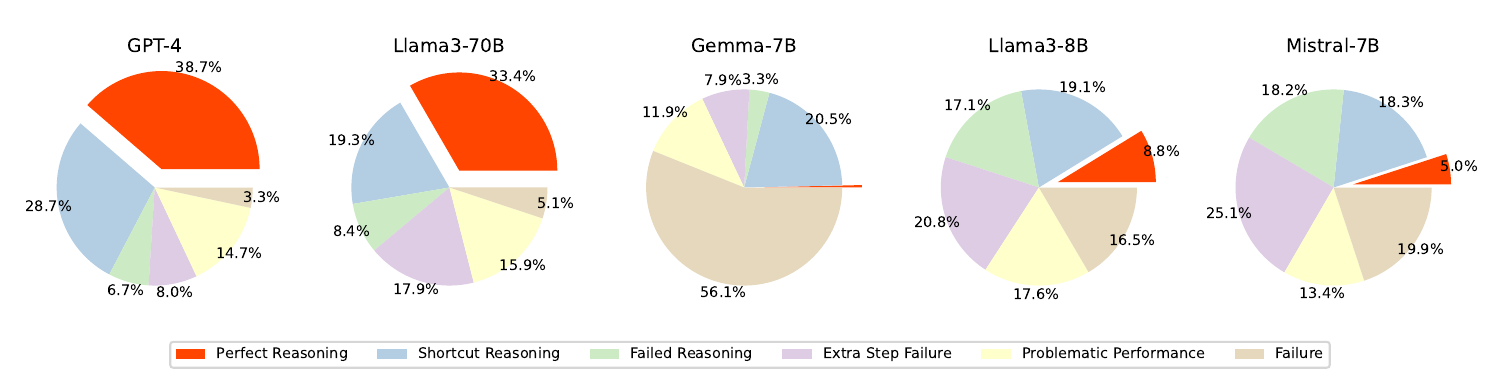}
    \caption{Distribution of performance categories of different LLMs on our dataset.}
    \label{fig:pie_charts}
\end{figure*}

\paragraph{Baseline Performance}
%
We observe that the performance of the baseline is low but non-zero, and better on the initial 2-hop questions (1.9 EM and 7.4 EM). 
As the scores are far from any other model's performance in both cases, this indicates that the models cannot directly use heuristics to solve most questions. 

\paragraph{Models vs. Human Performance}
As shown in \S\ref{sec_human_perform}, the average human performance and the human upper bound are 84.3 and 94.0, respectively. 
However, even in the best setting, GPT-4's performance is still lower than the average human score, indicating that there is room for improving the reasoning abilities of current models.

\paragraph{Our Question vs. Initial 2-hop Question}
Between the initial 2-hop questions (Case 2) and our new questions (Case 1), we observe a decrease in performance for both EM and F1 scores across all models when adding an additional hop, between up to 26.0 points in EM for GPT-4, to up to 53.8 points EM for Mistral-7B. Smaller models such as Mistral-7B and Llama3-8B seem to have a larger gap in performance between both cases compared to larger models. This indicates that our dataset is more challenging than the initial 2-hop datasets.

\paragraph{CoT Prompting}
All tested models benefit from the few shot-CoT prompting, gaining between 3.5 (Mistral-7B) and 23.0 (GPT-4) percentage points EM. 
The best performance is reached by GPT-4\_2-shot-cot prompting, which reaches 73.3 EM. Generally, larger models perform better, as both GPT-4 and Llama-70B reaching up to 73.3 and 59.2 EM, respectively, compared to between up to 11.3 and 30.5 EM for the models with 7-8 B parameters. 
During analysis, we observed that the result of Gemma-7B often refuses to answer. In our final results, we found from a total of 6,708 prompts, the answer contained the string ``I cannot answer'' up to 1,452 times (reached for 3-shot-cot).

\paragraph{Results on Six Cases}
As shown in the Figure, all models obtain high scores on the initial two-hop questions and its sub-questions (Case 2,5,6), but low scores on questions that include our added reasoning step (Case 1,3,4). 
It seems that our additional hop adds additional difficulty to the questions, apart from the fact that the questions get longer, since all models achieve higher scores on Case 5 and 6 compared to Case 4. We believe this is mainly due to the extractive answer type in Case 5 and 6. 
Similarly, when comparing Case 2 and Case 3, the models also achieve higher scores on Case 2 than on Case 3. 
In summary, our extended-hop approach increases the difficulty of the questions compared to the 2-hop extractive questions alone.


\subsection{Performance Category Analysis}
For a more detailed analysis of LLMs' performance, particularly the causes for the failures, we also ask the LLMs to answer the four other cases of the question, as shown in Figure~\ref{fig_overall}.  
We classify LLMs' performance into the 6 following categories based on whether they can correctly answer different cases. We also present the detailed categorization in Appendix~\ref{subsec:performance categorization}. 
\begin{itemize}[itemsep=0em, leftmargin=1.2em]
    \item \textbf{Perfect Reasoning}: the LLM answers all cases correctly.
    \item \textbf{Shortcut Reasoning}: the LLM answers the initial question correctly, but fails in either of its sub-questions. In this situation, it extracts the answer from the context instead of reasoning.
    \item \textbf{Failed Reasoning}: the LLM answers the sub-questions correctly but fails in the question.
    \item \textbf{Extra Step Failure}: the LLM fails to answer all the cases regarding our designed question from the template. In this situation, it is unable to perform the required type of reasoning. 
    \item \textbf{Problematic Performance}: the LLM answers the question correctly but inexplicably fails in some sub-questions, except shortcut reasoning. 
    \item \textbf{Failure}: other conditions.
\end{itemize}

Figure~\ref{fig:pie_charts} shows the distribution of performance categories of the LLMs on our dataset. All the models are prompted with 2-shot CoT examples because it shows the best overall performance across different models and cases. EM is the criterion used to determine whether the answer is correct or not. Consistent with the previous analysis, larger models (Llama3-70B and GPT-4) demonstrate more perfect reasoning compared with smaller models (Gemma-7B, Llama-8B, and Mistral-7B).  

Llama3-7B and GPT-4 exhibit different performance patterns. Only 8\% of extra step failure indicates that GPT-4 can better solve our designed template questions (Case 4) and their derivatives (Case 1, 3). For example, GPT-4 can correctly answer most questions in the format of \textit{How many repeated letters are there in the first name of \#Name?}, while Llama3-70B fails in some of these questions. It turns out Llama3 does not conduct arithmetic reasoning, commonsense reasoning and symbolic reasoning so well as GPT-4.

However, GPT-4 faces a substantial issue with shortcut reasoning. In 28.7\% of the questions, GPT-4 can correctly answer the initial 2-hop question (Case 2) but fails in either of its sub-questions (Case 5 and Case 6). 
In contrast, Llama3-70B shows a ``Shortcut Reasoning'' rate of 19.3\%.
Thus, despite GPT-4's strong overall performance, our findings suggest that it heavily relies on shortcut reasoning to answer multi-hop questions. 
This highlights the need for a more detailed analysis when comparing the reasoning capabilities of different models




\section{Conclusion}
We introduce a new multi-hop dataset by extending existing 2-hop datasets with an additional hop. 
A notable aspect is that, through careful template design and selection of 2-hop samples, we transition from extractive to generative answers. 
Additionally, our samples require various types of reasoning to address the questions. 
Human performance scores indicate that our dataset is of high quality and suitable for evaluating models. 
We then use our dataset to evaluate the reasoning capabilities of five LLMs.
Experimental results reveal a large gap between LLMs and human performance. 
Our analyses further demonstrate that the generative questions in our dataset are challenging for the models, preventing them from relying on simple heuristics to extract answers from the provided paragraphs.

\section*{Ethical Statement and Broader Impact}
Our dataset builds upon publicly available datasets, which themselves use publicly available information. The users were not asked to provide any information, and explicitly asked the users to fulfill a very narrow task, that did especially involve using only the available information. 
Human annotators were volunteer students on the Master's and PhD levels and professors working on research in an NLP Lab, who were given the opportunity to propose and execute their own annotation task with the same group of annotators in return. The annotators received an in-depth introduction including the topic of the research, and details about the intended use of the dataset.

Our work could help the community to benchmark new models and understand whether models are able to perform reasoning, an important next step in the development of intelligent models.

\section*{Limitations}
There are three limitations in our study. 
The first one concerns the diversity of the dataset. 
Although we try to use the three existing multi-hop datasets, our extended-hop questions are derived from designed templates (about 97 templates), which are not as diverse as non-template questions.
The second point concerns our generated answers. These answers are not fully verified, as they are produced via code, based on the initial 2-hop answers. While we manually check the answers for all templates, we only verify a few samples per template, meaning not all answers are thoroughly reviewed. If unexpected cases occur that are not handled by our code, this may result in incorrect answers.
The third point concerns running GPT-4. We have 6 settings per model, each with 6 cases (different types of questions), resulting in 36 runs per sample for one model. Due to the cost, we only ran GPT-4 on 150 samples.

\section*{Acknowledgments}
We would like to thank Jonas Lührs, Juan Junqueras, Kon Woo Kim, Léane Jourdan, and Tomás Vergara Browne for joining our dataset annotation tasks.
This work was supported by JSPS KAKENHI Grant Numbers 24K03231 and 22K17954  and JST PRESTO Grant Number JPMJPR20C4.

\bibliography{anthology,custom}

\appendix

\section{Dataset Creation Process}
\label{appendix_data_sec}

\subsection{Licenses}
HotpotQA and MusiQue were published under the CC BY-SA 4.0 license, which explicitly allows adaptation. 2WikiMultihopQA was published under the Apache License 2.0, which also allows for distribution and modification. We intend to publish our newly generated dataset under the CC BY-SA 4.0 license.

\subsection{MuSiQue Dataset}
\label{appendix_data_sec_musi}

We present three examples: (1) issues with disconnected reasoning, (2) lack of evidence to support the answer, and (3) multiple answers arising from setting questions without using the provided paragraphs in Tables \ref{tab_disconnected_musi}, \ref{tab_no_evidence_musi}, and \ref{tab_multiple_ans_musi}, respectively.

\begin{table*}[h]
\centering
 \resizebox{\textwidth}{!}{
\begin{tabular}{l p{14cm}}
\toprule

 \textbf{id:} 2hop\_\_752214\_639679 \\ 
 
\multirow{1}{14cm}{\textbf{question:} Who is the spouse of the author of Queen of the Elephants?} \\ 

\textbf{answer:} Clio Goldsmith \\

\multirow{7}{14cm}{\textbf{question\_decomposition:} \\
           - sub question 1: Queen of the Elephants >> author \\
           - sub answer 1: Mark Shand \\
           - sub paragraph\_support\_title 1: Queen of the Elephants \\
   
           - sub question 2: \#1 >> spouse \\ 
           - sub answer 2: Clio Goldsmith \\
           - sub paragraph\_support\_title 2: Clio Goldsmith \\
        
         } \\ \\ \\ \\ \\ \\ \\ 
         
\multirow{10}{14cm}{\textbf{context:} \\
\textit{Paragraph 1: Queen of the Elephants} \\
Queen of the Elephants is a book written by the conservationist and travel writer Mark Shand and the corresponding BBC documentary \"Queen of the Elephants\", based on the life of the first female mahout in recent times--Parbati Barua of Kaziranga. The book went on to win the award, providing free publicity simultaneously to the profession of mahouts, and to Kaziranga. \\ 
\textit{Paragraph 2: Clio Goldsmith} \\ 
Clio Goldsmith (born 16 June 1957) is a French former actress, appearing mostly as a Femme fatale in some films of the early 1980s. She is a member of the prominent Goldsmith family through her father ecologist Edward Goldsmith.
} \\  \\ \\ \\ \\\\ \\\\  \\ \\ \\

\bottomrule
\end{tabular}}
\caption{
This is an example of disconnected reasoning in MuSiQue: as shown in this example, from the answer of the first sub-question (Mark Shand), we have no evidence to proceed to the final answer (Clio Goldsmith).
}
\label{tab_disconnected_musi}
\end{table*}

\begin{table*}[h]
\centering
 \resizebox{\textwidth}{!}{
\begin{tabular}{l p{14cm}}
\toprule

 \textbf{id:} 2hop\_\_623931\_656446 \\ 
 
\multirow{1}{14cm}{\textbf{question:} Who is the spouse of a cast member of Secrets of a Windmill Girl?} \\ 

\textbf{answer:} John Alderton \\

\multirow{7}{14cm}{\textbf{question\_decomposition:} \\
           - sub question 1: Secrets of a Windmill Girl >> cast member \\
           - sub answer 1: Pauline Collins \\
           - sub paragraph\_support\_title 1: Secrets of a Windmill Girl \\
   
           - sub question 2: \#1 >> spouse \\ 
           - sub answer 2: John Alderton \\
           - sub paragraph\_support\_title 2: Mrs Caldicot's Cabbage War \\
        
         } \\ \\ \\ \\ \\ \\ \\ 
         
\multirow{10}{14cm}{\textbf{context:} \\
\textit{Paragraph 1: Secrets of a Windmill Girl} \\
Secrets of a Windmill Girl is a 1966 British exploitation film directed by Arnold L Miller. It recounts the road to ruin of a young woman (Pauline Collins) who becomes involved with the striptease scene after becoming a dancer at the Windmill Theatre in London. The film features fan dances by former Windmill Theatre Company performers. It was originally released in Britain as part of a double bill with \"Naked as Nature Intended\". \\ 
\textit{Paragraph 2: Mrs Caldicot's Cabbage War} \\ 
Mrs Caldicot's Cabbage War is a British comedy-drama film from 2002, directed by Ian Sharp and starring Pauline Collins, John Alderton and Peter Capaldi. It is based on a 1993 novel with the same name by Vernon Coleman.
} \\  \\ \\ \\ \\\\ \\\\  \\ \\ \\

\bottomrule
\end{tabular}}
\caption{
This is an example in MuSiQue where we do not have enough evidence to infer that the final answer (the spouse of  Pauline Collins) is John Alderton.
}
\label{tab_no_evidence_musi}
\end{table*}

\begin{table*}[h]
\centering
 \resizebox{\textwidth}{!}{
\begin{tabular}{l p{14cm}}
\toprule

 \textbf{id:} 2hop\_\_252311\_366220 \\ 
 
\multirow{1}{14cm}{\textbf{question:} Who founded the company that distributed the film UHF?} \\ 

\textbf{answer:} Mike Medavoy \\

\multirow{7}{14cm}{\textbf{question\_decomposition:} \\
           - sub question 1: UHF >> distributed by \\
           - sub answer 1: Orion Pictures \\
           - sub paragraph\_support\_title 1: UHF (film) \\
   
           - sub question 2: \#1 >> founded by \\ 
           - sub answer 2: Mike Medavoy \\
           - sub paragraph\_support\_title 2: Mike Medavoy \\
        
         } \\ \\ \\ \\ \\ \\ \\ 
         
\multirow{13}{14cm}{\textbf{context:} \\
\textit{Paragraph 1: UHF (film)} \\
Yankovic and Levey wrote the film after Yankovic's second studio album, looking to apply the musician's parody and comedy to film, and chose the approach of George being a straight man with a vivid imagination to support the inclusion of parodies within the film. They struggled with finding a film production company for financing the film, but were eventually able to get Orion Pictures' support after stating they could keep the film costs under \$5 million. Principal filming took place around Tulsa, Oklahoma, with many of the extras for the film from the Tulsa and Dallas, Texas areas. \\ 
\textit{Paragraph 2: Mike Medavoy} \\ 
Morris Mike Medavoy (born January 21, 1941) is an American film producer and executive, co-founder of Orion Pictures (1978), former chairman of TriStar Pictures, former head of production for United Artists (1974\u20131978) and current chairman and CEO of Phoenix Pictures.
} \\  \\ \\ \\ \\\\ \\\\  \\ \\ \\  \\ \\ \\

\bottomrule
\end{tabular}}
\caption{
This is an example in MuSiQue. If we use the two provided paragraphs, the answer to the question is Mike Medavoy. However, if we do not use these paragraphs, there are multiple possible answers to the question because the Orion Pictures company was founded by five people: Arthur B. Krim, Eric Pleskow, Mike Medavoy, William Bernstein, and Robert Benjamin.
}
\label{tab_multiple_ans_musi}
\end{table*}

\subsection{Dataset generation details}

We make use of various libraries to generate the answers to our dataset. For questions regarding the number of syllables, we make use of NLTK and use cmudict to estimate this number. 
To deal with place answers, we use the Nominatim API to search for places on OpenStreetView and retrieve the coordinates for each place mentioned in earlier datasets.

\subsection{Human Verification} 
\label{appendix_data_sec_verification}
We provide the following guidelines to annotators during the annotation process.

\begin{itemize}
    \item Check the questions with \textit{New Question (Overall)} or \textit{New Question (Sub-question)} labels.

    \item If a question is good, give it an [\textbf{OK}] label.

    \item If a question is understandable but has some flaws (e.g., grammar, typo, etc.), give it a [\textbf{Modified}] label and please correct it.

    \item If a question is not understandable at all, give it an [\textbf{Issue}] label and briefly explain which part is confusing in the comment cell.

    \item Three additional fields are provided as \textbf{Reference}: \textit{New Answer}, \textit{Original Question}, and \textit{Original Answer}. You don’t need to check the correctness. However, if you find any severe issues (e.g., difficult to understand, the answer doesn’t address the question, or messy code), please add a comment in the corresponding rows.
\end{itemize}

\begin{figure*}[ht]
\centering
    \includegraphics[scale=0.45]{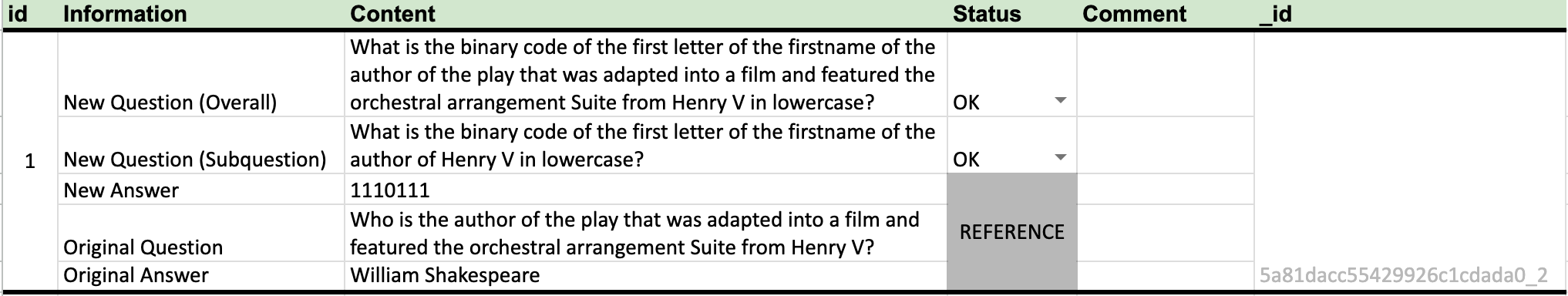}
    \caption{
    Our annotation interface.
    }
    \label{fig_annotate_interface}
\end{figure*}

Figure \ref{fig_annotate_interface} shows our annotation interface.
We also provide the explanations for each field in the annotation guideline:

\begin{itemize}
    \item \textit{New Question (Overall)}: our new question  
\item \textit{New Question (Sub-question)}: our new question but we only put the top question on the second hop. (in \textit{New Question (Overall)}, we put the top question on the full 2-hop question)

\item \textit{New Answer}: an answer for a New Question (Overall)

\item \textit{Original Question}: the initial 2-hop question
\item \textit{Original Answer}: the answer for the Original Question
\end{itemize}

\subsection{Our Dataset Information}
\label{appendix_data_sec_our_data}
Each sample in our dataset contains the following information:

\begin{itemize}
    \item \_id: a unique id for each sample
    \item question: our new question
    \item answer: our new answer
    \item previous\_question: the previous 2-hop question
    \item previous\_answer: the previous 2-hop answer
    \item question\_decomposition: a list of sub-questions and sub-answers
    \item context: the two gold paragraphs
    \item answer\_type: an answer type of the new question
    \item previous\_answer\_type: the answer type of the previous 2-hop question
    \item no\_of\_hops: the number of hops in our extended question
    \item reasoning\_type: the list of required reasoning types
    \item pattern: a template that is used to generate the new question
    \item subquestion\_patterns: a list of sub-questions of the template that is used to generate the new question 
    \item cutted\_question: the noun form that we obtain from the previous 2-hop question
    \item ques\_on\_last\_hop: instead of integrating the new hop into the entire previous 2-hop question, we integrate it into the second hop of the previous 2-hop question.
    This is the third case (Case 3) in Figure \ref{fig_overall}.
    
\end{itemize}

We present an example in our dataset in Table \ref{tab_example}.

\begin{table*}[h]
\centering
 \resizebox{\textwidth}{!}{
\begin{tabular}{l p{14cm}}
\toprule

 \textbf{\_id:} fc0370920baf11ebab90acde48001122\_14 \\ 
 
\multirow{2}{14cm}{\textbf{question:} What is the concatenation of the last letter of the first name and the first letter of the last name of the paternal grandmother of Mervyn Tuchet, 4Th Earl Of Castlehaven in lowercase?} \\ \\ \\

\textbf{answer:} ym \\ 

\textbf{previous\_question:} Who is the paternal grandmother of Mervyn Tuchet, 4Th Earl Of Castlehaven? \\ 

\textbf{previous\_answer:} Lucy Mervyn \\ 

\multirow{14}{14cm}{\textbf{question\_decomposition:} \\
           - sub question 1: Who is the father of Mervyn Tuchet, 4Th Earl Of Castlehaven? \\
           - sub answer 1: Mervyn Tuchet, 2nd Earl of Castlehaven \\
           - sub paragraph\_support\_title 1: Mervyn Tuchet, 4th Earl of Castlehaven \\
   
           - sub question 2: Who is the mother of Mervyn Tuchet, 2Nd Earl Of Castlehaven? \\ 
           - sub answer 2: Lucy Mervyn \\
           - sub paragraph\_support\_title 2: Mervyn Tuchet, 2nd Earl of Castlehaven \\
           - sub question 3: What is the concatenation of the last letter of the first name and the first letter of the lastname of Lucy Mervyn in lowercase? \\ 
           - sub answer 3: ym \\ 
           - sub paragraph\_support\_title 3: \\   
           - details: the details for the third sub-question \\ 
         } \\ \\ \\ \\ \\ \\ \\ \\ \\ \\ \\ \\ \\
         
\multirow{11}{14cm}{\textbf{context:} \\
\textit{Paragraph 1: Mervyn Tuchet, 4th Earl of Castlehaven} \\
Mervyn Tuchet, 4th Earl of Castlehaven (died 2 November 1686) was the third son of Mervyn Tuchet, 2nd Earl of Castlehaven, and his first wife, Elizabeth Barnham (1592 – c. 1622).,
He married Mary Talbot (buried 15 March 1710/1), daughter of John Talbot, 10th Earl of Shrewsbury (bef.,1601–1654) and his wife, née Mary Fortesque., ... \\ 
\textit{Paragraph 2: Mervyn Tuchet, 2nd Earl of Castlehaven} \\ 
Mervyn Tuchet (sometimes Mervin Touchet), 2nd Earl of Castlehaven (1593 – 14 May 1631), was an English nobleman who was convicted of rape and sodomy and subsequently executed.,
A son of George Tuchet, 1st Earl of Castlehaven and 11th Baron Audley, by his wife, Lucy Mervyn, he was known by the courtesy title of Lord Audley during his father's lifetime, so is sometimes referred to as Mervyn Audley., ...
} \\  \\ \\ \\ \\\\ \\\\  \\ \\ \\ \\

\textbf{answer\_type:} string \\ 

\textbf{previous\_answer\_type:} person \\

\textbf{no\_of\_hops:} 5 \\ 

\textbf{reasoning\_type:} Symbolic, Commonsense \\ 

\multirow{2}{14cm}{\textbf{pattern:} What is the concatenation of the last letter of the first name and the first letter of the last name of \#Name in lowercase?} \\ \\

\multirow{5}{14cm}{\textbf{subquestion\_patterns:} \\ 
         What is the first name of \#Name? \\
         What is the last letter of \#Ans1? \\
         What is the last name of \#Name? \\ 
         What is the first letter of \#Ans3? \\
         What is the concatenation of \#Ans2 and \#Ans4?}
\\ \\ \\ \\ \\ \\

\textbf{cutted\_question:} the paternal grandmother of Mervyn Tuchet, 4Th Earl Of Castlehaven \\

\multirow{3}{14cm}{\textbf{ques\_on\_last\_hop:} What is the concatenation of the last letter of the first name and the first letter of the lastname of the mother of Mervyn Tuchet, 2Nd Earl Of Castlehaven in lowercase?} \\ \\ \\

\bottomrule
\end{tabular}}
\caption{
An example containing all information in our dataset.
Due to the space limitation, we present the field `details' in the `question decomposition' part in Table~\ref{tab_app_details}.
}
\label{tab_example}
\end{table*}

\begin{table}[h]
\centering
 \resizebox{\columnwidth}{!}{
\begin{tabular}{l p{14cm}}
\toprule

 \textbf{sub\_id:} 3\_1 \\ 
\textbf{question:} What is the first name of Lucy Mervyn?\\ 
\textbf{answer:} Lucy \\ \midrule

 \textbf{sub\_id:} 3\_2 \\ 
\textbf{question:} What is the last letter of Lucy? \\ 
\textbf{answer:} y \\ \midrule

 \textbf{sub\_id:} 3\_3 \\ 
\textbf{question:} What is the last name of Lucy Mervyn?\\ 
\textbf{answer:} Mervyn \\ \midrule

 \textbf{sub\_id:} 3\_4 \\ 
\textbf{question:} What is the first letter of Mervyn?\\ 
\textbf{answer:} m \\ \midrule

 \textbf{sub\_id:} 3\_5 \\ 
\textbf{question:} What is the concatenation of y and m?\\ 
\textbf{answer:} ym \\ 

\bottomrule
\end{tabular}}
\caption{
Example of the field `details' in the `question decomposition' part in Table~\ref{tab_example}.
}
\label{tab_app_details}
\end{table}

\subsection{Dataset Analysis}
\label{app_data_analysis}

As mentioned in Section \ref{sec_dataset_creation}, there are four answer types in our dataset: date, number, string, and letter.
We present examples for each type of answer in Table \ref{app_tab_ans_type}.

Each sample in our dataset includes a list of question decompositions that can be useful for detailed analysis of the results. In addition, we include Case 3 (as shown in Figure~\ref{fig_overall}), where we extend the second hop of the previous 2-hop question, rather than extending the entire previous 2-hop question. Currently, we use numbers to differentiate between these cases. The explanation for each case is as follows:

\begin{itemize}
    \item Case 1: Our newly generated question
    \item Case 2: The previous 2-hop question
    \item Case 3: Our newly 2-hop generated question
    \item Case 4: Our extended question 
    \item Case 5: The second hop of the previous 2-hop question
    \item Case 6: The first hop of the previous 2-hop question
\end{itemize}

In MoreHopQA {\small w/ hv}, we also ask humans to verify Case 3.

For 2Wiki and MuSiQue, the questions in Case 3 are automatically created using the same process as for questions in Case 1. In HotpotQA, to enhance efficiency, we use GPT-4 as the annotator to create the questions in Case 3.

\begin{table*}[ht]
\centering
 \resizebox{\textwidth}{!}{
\begin{tabular}{l l l p{12.5cm} p{1.7cm} p{1.0cm}}

\toprule

\textbf{Question} & \textbf{Answer} & \textbf{Type} \\

\midrule

What is the date one day after when Prince Nikolai Of Denmark's mother was born? & 1964-07-01 & Date \\ \midrule

\multirow{2}{12.5cm}{How many letters are there between the first and last letters of the first name of the director of a 2004 film where Kam Heskin plays Paige Morgan in?} & \multirow{2}{1.5cm}{4} & \multirow{2}{1.5cm}{Number} \\ \\ \midrule

\multirow{2}{12.5cm}{What is the alphabetical order of the letters in the last name of the father of the director of film My 20Th Century?} & \multirow{2}{1.5cm}{deeiny} & \multirow{2}{1.5cm}{String} \\ \\ \midrule

What is the last letter of the last name of the father of Empress Wang's husband? & i & Letter \\

\bottomrule
\end{tabular} }
\caption{
Examples of different answer types in our dataset. 
}
\label{app_tab_ans_type}
\end{table*}

\section{Experiments}
\label{app_experiment}

\subsection{Experimental Details}
We run Llama-3-8B-Instruct, Mistral-7B-Instruct-v0.3 and Gemma-7B-it on a single GPU (NVIDIA A100 40 GB), and Llama-3-70B-Instruct on 2 NVIDIA A100 80 GB GPUs. 
We use the following decoding parameters for all models: do\_sample=True, max\_new\_tokens=256. 
The entire experiments took a total of 18 hours of runtime on the single GPU, and 30 hours on the pair of GPUs for LLama-3-70B. We additionally spent 84 \$ to run GPT-4-Turbo.
We wrote the Code for Evaluation with the help of Github Copilot.

For NER in the postprocessing of the model answers as described in section 5.1, we used the NER module from spacy's en\_core\_web\_sm pipeline. Please also see our published code for more details. 

\subsection{Results}
\label{app_experiment_result}

The full results are presented in Table~\ref{tab_main_result}.

\begin{table*}[ht]
\begin{center}
\resizebox{\textwidth}{!}{%
        \begin{tabular}{l rr rr rr rr rr rr}
    \toprule

            \multirow{2}{*}{\textbf{Model}} & 
        \multicolumn{2}{c}{\textbf{Case 1}} & \multicolumn{2}{c}{\textbf{Case 2}}  & \multicolumn{2}{c}{\textbf{Case 3}}  & \multicolumn{2}{c}{\textbf{Case 4}}  & \multicolumn{2}{c}{\textbf{Case 5}}  & \multicolumn{2}{c}{\textbf{Case 6}} \\
        \cmidrule(rl){2-3}
    \cmidrule(rl){4-5}
    \cmidrule(rl){6-7}
    \cmidrule(rl){8-9}
    \cmidrule(rl){10-11}
    \cmidrule(rl){12-13}
         ~ & EM & F1 & EM & F1 & EM & F1 & EM & F1 & EM & F1 & EM & F1\\   \midrule

Baseline\_zero-cot &1.88$_{\pm0.81}$ & 7.93$_{\pm1.28}$ & 7.42$_{\pm1.70}$ & 20.12$_{\pm1.90}$ &  &  &  &  &  &  &  &  \\ \midrule
Llama-8B\_zeroshot & 23.26$_{\pm2.50}$ & 28.62$_{\pm2.50}$ & 66.99$_{\pm2.86}$ & 79.82$_{\pm2.01}$ & 26.03$_{\pm2.50}$ & 30.38$_{\pm2.59}$ & 35.69$_{\pm3.04}$ & 37.62$_{\pm2.86}$ & 85.33$_{\pm2.15}$ & 91.68$_{\pm1.35}$ & 75.31$_{\pm2.50}$ & 87.36$_{\pm1.61}$  \\ 
Llama-8B\_2-shot & 24.06$_{\pm2.42}$ & 28.54$_{\pm2.45}$ & 66.91$_{\pm2.68}$ & 77.79$_{\pm2.18}$ & 28.18$_{\pm2.50}$ & 32.14$_{\pm2.56}$ & 35.69$_{\pm2.68}$ & 37.52$_{\pm2.54}$ & 86.31$_{\pm2.06}$ & 92.53$_{\pm1.25}$ & 78.26$_{\pm2.33}$ & 88.51$_{\pm1.56}$  \\ 
Llama-8B\_3-shot &23.17$_{\pm2.42}$ & 27.74$_{\pm2.54}$ & 71.20$_{\pm2.77}$ & 80.45$_{\pm2.03}$ & 28.35$_{\pm2.86}$ & 32.22$_{\pm2.71}$ & 37.66$_{\pm2.95}$ & 39.24$_{\pm3.01}$ & 86.94$_{\pm1.97}$ & 92.66$_{\pm1.24}$ & 78.18$_{\pm2.42}$ & 88.56$_{\pm1.55}$  \\
Llama-8B\_zero-cot & 20.84$_{\pm2.59}$ & 26.48$_{\pm2.56}$ & 65.03$_{\pm2.68}$ & 78.07$_{\pm1.96}$ & 24.87$_{\pm2.68}$ & 29.38$_{\pm2.67}$ & 34.08$_{\pm2.86}$ & 36.36$_{\pm2.90}$ & 83.99$_{\pm2.33}$ & 90.69$_{\pm1.47}$ & 74.42$_{\pm2.50}$ & 86.76$_{\pm1.60}$  \\
Llama-8B\_2-shot-cot & 28.26$_{\pm2.77}$ & 32.28$_{\pm2.65}$ & 69.14$_{\pm2.68}$ & 79.72$_{\pm1.99}$ & 34.08$_{\pm2.95}$ & 37.69$_{\pm2.78}$ & 45.97$_{\pm3.04}$ & 47.85$_{\pm2.81}$ & 85.15$_{\pm2.06}$ & 91.80$_{\pm1.39}$ & 77.82$_{\pm2.50}$ & 88.04$_{\pm1.52}$  \\
Llama-8B\_3-shot-cot & 30.50$_{\pm2.59}$ & 34.38$_{\pm2.70}$ & 73.26$_{\pm2.59}$ & 82.07$_{\pm1.95}$ & 34.44$_{\pm2.86}$ & 38.12$_{\pm2.74}$ & 45.71$_{\pm2.95}$ & 47.26$_{\pm2.90}$ & 86.31$_{\pm2.15}$ & 92.16$_{\pm1.34}$ & 76.48$_{\pm2.42}$ & 86.26$_{\pm1.91}$  \\ \midrule

Mistral-7B\_zeroshot & 14.49$_{\pm2.06}$ & 20.87$_{\pm2.15}$ & 64.04$_{\pm2.77}$ & 73.85$_{\pm2.33}$ & 18.96$_{\pm2.42}$ & 24.36$_{\pm2.31}$ & 27.73$_{\pm2.68}$ & 30.59$_{\pm2.59}$ & 77.28$_{\pm2.50}$ & 83.13$_{\pm1.98}$ & 65.21$_{\pm2.68}$ & 78.72$_{\pm1.97}$  \\
Mistral-7B\_2-shot & 17.17$_{\pm2.42}$ & 23.52$_{\pm2.42}$ & 69.68$_{\pm2.95}$ & 78.17$_{\pm2.26}$ & 22.90$_{\pm2.59}$ & 28.09$_{\pm2.59}$ & 33.72$_{\pm2.86}$ & 35.96$_{\pm2.81}$ & 84.53$_{\pm2.15}$ & 89.80$_{\pm1.68}$ & 76.39$_{\pm2.59}$ & 86.32$_{\pm1.70}$  \\
Mistral-7B\_3-shot & 16.73$_{\pm2.15}$ & 23.17$_{\pm2.37}$ & 70.57$_{\pm2.86}$ & 78.37$_{\pm2.32}$ & 23.52$_{\pm2.50}$ & 28.40$_{\pm2.52}$ & 32.74$_{\pm2.95}$ & 35.17$_{\pm2.85}$ & 84.35$_{\pm2.24}$ & 89.92$_{\pm1.72}$ & 76.65$_{\pm2.59}$ & 86.29$_{\pm1.67}$  \\
Mistral-7B\_zero-cot & 18.16$_{\pm2.24}$ & 23.94$_{\pm2.37}$ & 55.64$_{\pm2.86}$ & 68.33$_{\pm2.16}$ & 20.04$_{\pm2.50}$ & 25.40$_{\pm2.38}$ & 30.59$_{\pm2.77}$ & 33.90$_{\pm2.64}$ & 66.82$_{\pm2.95}$ & 77.15$_{\pm2.18}$ & 50.18$_{\pm3.04}$ & 70.78$_{\pm2.18}$  \\
Mistral-7B\_2-shot-cot & 17.80$_{\pm2.33}$ & 23.88$_{\pm2.46}$ & 68.96$_{\pm2.68}$ & 77.48$_{\pm2.14}$ & 24.87$_{\pm2.59}$ & 29.97$_{\pm2.61}$ & 37.48$_{\pm2.86}$ & 40.15$_{\pm2.91}$ & 85.51$_{\pm2.15}$ & 90.76$_{\pm1.55}$ & 75.85$_{\pm2.59}$ & 85.64$_{\pm1.98}$  \\
Mistral-7B\_3-shot-cot & 19.41$_{\pm2.42}$ & 25.75$_{\pm2.44}$ & 68.34$_{\pm2.77}$ & 76.92$_{\pm2.19}$ & 25.94$_{\pm2.59}$ & 31.12$_{\pm2.67}$ & 37.57$_{\pm2.77}$ & 40.15$_{\pm2.84}$ & 85.69$_{\pm2.15}$ & 91.00$_{\pm1.52}$ & 75.49$_{\pm2.59}$ & 85.69$_{\pm1.80}$  \\ \midrule

Gemma-7B\_zeroshot & 7.07$_{\pm1.52}$ & 12.81$_{\pm1.78}$ & 40.07$_{\pm3.04}$ & 49.24$_{\pm2.62}$ & 9.48$_{\pm1.79}$ & 14.77$_{\pm1.82}$ & 18.87$_{\pm2.42}$ & 24.52$_{\pm2.32}$ & 59.12$_{\pm2.95}$ & 69.91$_{\pm2.35}$ & 52.86$_{\pm2.86}$ & 69.78$_{\pm2.15}$  \\
Gemma-7B\_2-shot & 11.27$_{\pm1.88}$ & 16.85$_{\pm2.04}$ & 32.83$_{\pm2.86}$ & 41.09$_{\pm2.55}$ & 13.15$_{\pm1.97}$ & 18.11$_{\pm2.15}$ & 21.74$_{\pm2.59}$ & 26.31$_{\pm2.56}$ & 50.89$_{\pm3.04}$ & 61.64$_{\pm2.46}$ & 40.52$_{\pm2.95}$ & 57.63$_{\pm2.31}$  \\
Gemma-7B\_3-shot & 8.94$_{\pm1.70}$ & 14.71$_{\pm1.81}$ & 27.91$_{\pm2.68}$ & 37.41$_{\pm2.55}$ & 12.52$_{\pm2.06}$ & 17.76$_{\pm2.04}$ & 22.09$_{\pm2.59}$ & 26.63$_{\pm2.68}$ & 44.99$_{\pm3.04}$ & 55.99$_{\pm2.62}$ & 34.70$_{\pm2.77}$ & 52.00$_{\pm2.27}$  \\
Gemma-7B\_zero-cot & 7.33$_{\pm1.52}$ & 13.23$_{\pm1.73}$ & 33.81$_{\pm2.86}$ & 43.80$_{\pm2.62}$ & 10.02$_{\pm1.79}$ & 15.73$_{\pm1.86}$ & 15.74$_{\pm2.24}$ & 21.81$_{\pm2.26}$ & 49.73$_{\pm3.13}$ & 61.99$_{\pm2.58}$ & 39.53$_{\pm2.86}$ & 58.44$_{\pm2.29}$  \\
Gemma-7B\_2-shot-cot & 10.55$_{\pm1.97}$ & 15.46$_{\pm1.84}$ & 31.57$_{\pm2.59}$ & 39.76$_{\pm2.52}$ & 13.51$_{\pm2.06}$ & 18.00$_{\pm2.18}$ & 10.91$_{\pm1.88}$ & 16.62$_{\pm1.97}$ & 59.48$_{\pm3.13}$ & 69.11$_{\pm2.63}$ & 42.84$_{\pm3.04}$ & 59.71$_{\pm2.34}$  \\
Gemma-7B\_3-shot-cot & 10.82$_{\pm1.88}$ & 15.04$_{\pm1.91}$ & 24.15$_{\pm2.50}$ & 32.85$_{\pm2.44}$ & 11.90$_{\pm1.97}$ & 16.92$_{\pm2.16}$ & 11.99$_{\pm1.97}$ & 16.54$_{\pm2.06}$ & 56.53$_{\pm3.13}$ & 66.84$_{\pm2.44}$ & 40.34$_{\pm2.86}$ & 56.83$_{\pm2.43}$ \\ \midrule

GPT-4\_zeroshot & 51.33$_{\pm8.67}$ & 53.11$_{\pm8.33}$ & 87.33$_{\pm6.00}$ & 91.29$_{\pm4.22}$ & 64.00$_{\pm8.00}$ & 65.78$_{\pm7.78}$ & 67.33$_{\pm8.00}$ & 67.67$_{\pm7.67}$ & 90.00$_{\pm5.33}$ & 92.81$_{\pm4.30}$ & 70.67$_{\pm7.33}$ & 83.86$_{\pm4.93}$  \\
GPT-4\_2-shot & 72.00$_{\pm7.33}$ & 73.44$_{\pm7.00}$ & 88.00$_{\pm6.00}$ & 90.91$_{\pm4.93}$ & 72.00$_{\pm7.33}$ & 74.11$_{\pm6.67}$ & 74.00$_{\pm7.33}$ & 74.33$_{\pm7.33}$ & 90.67$_{\pm5.33}$ & 92.65$_{\pm4.38}$ & 74.67$_{\pm7.33}$ & 86.08$_{\pm4.89}$  \\
GPT-4\_3-shot & 68.67$_{\pm7.33}$ & 70.11$_{\pm7.44}$ & 88.67$_{\pm5.33}$ & 91.05$_{\pm4.49}$ & 69.33$_{\pm7.33}$ & 70.80$_{\pm7.20}$ & 80.00$_{\pm6.67}$ & 80.00$_{\pm6.67}$ & 86.67$_{\pm5.33}$ & 89.31$_{\pm4.49}$ & 76.00$_{\pm6.67}$ & 86.26$_{\pm4.97}$  \\
GPT-4\_zero-cot & 72.67$_{\pm7.33}$ & 72.70$_{\pm7.30}$ & 88.00$_{\pm5.33}$ & 91.69$_{\pm4.33}$ & 74.67$_{\pm7.33}$ & 76.67$_{\pm6.67}$ & 81.33$_{\pm6.67}$ & 81.33$_{\pm6.67}$ & 90.00$_{\pm4.67}$ & 92.43$_{\pm3.91}$ & 67.33$_{\pm8.00}$ & 81.51$_{\pm5.18}$  \\
GPT-4\_2-shot-cot & 73.33$_{\pm7.33}$ & 74.44$_{\pm7.11}$ & 88.67$_{\pm5.33}$ & 92.32$_{\pm3.90}$ & 77.33$_{\pm7.33}$ & 79.02$_{\pm6.62}$ & 82.00$_{\pm7.33}$ & 81.67$_{\pm7.00}$ & 90.00$_{\pm4.67}$ & 91.98$_{\pm4.27}$ & 72.67$_{\pm7.33}$ & 83.95$_{\pm4.88}$  \\
GPT-4\_3-shot-cot & 72.00$_{\pm7.33}$ & 73.13$_{\pm6.87}$ & 90.67$_{\pm5.33}$ & 93.54$_{\pm4.16}$ & 75.33$_{\pm6.67}$ & 76.78$_{\pm6.69}$ & 78.00$_{\pm6.67}$ & 78.42$_{\pm6.60}$ & 90.67$_{\pm4.67}$ & 93.09$_{\pm3.93}$ & 72.00$_{\pm7.33}$ & 84.71$_{\pm5.01}$ \\ \midrule

Llama-70B\_zeroshot & 36.23$_{\pm3.04}$ & 38.46$_{\pm3.00}$ & 82.56$_{\pm2.33}$ & 90.18$_{\pm1.43}$ & 39.18$_{\pm3.04}$ & 41.06$_{\pm2.95}$ & 50.63$_{\pm3.13}$ & 50.79$_{\pm3.16}$ & 87.75$_{\pm1.97}$ & 92.96$_{\pm1.25}$ & 77.55$_{\pm2.50}$ & 88.03$_{\pm1.58}$  \\
Llama-70B\_2-shot & 38.10$_{\pm2.86}$ & 39.80$_{\pm2.82}$ & 85.69$_{\pm2.06}$ & 92.23$_{\pm1.26}$ & 41.86$_{\pm2.95}$ & 43.43$_{\pm2.86}$ & 52.68$_{\pm3.13}$ & 52.77$_{\pm3.22}$ & 89.53$_{\pm1.88}$ & 93.89$_{\pm1.30}$ & 80.41$_{\pm2.42}$ & 89.59$_{\pm1.55}$  \\
Llama-70B\_3-shot & 38.64$_{\pm2.95}$ & 40.30$_{\pm3.04}$ & 87.21$_{\pm1.97}$ & 92.96$_{\pm1.34}$ & 40.97$_{\pm2.86}$ & 42.40$_{\pm2.91}$ & 52.24$_{\pm3.31}$ & 52.59$_{\pm3.26}$ & 89.71$_{\pm1.88}$ & 94.14$_{\pm1.30}$ & 80.50$_{\pm2.33}$ & 89.76$_{\pm1.56}$  \\
Llama-70B\_zero-cot & 49.91$_{\pm2.95}$ & 51.29$_{\pm3.06}$ & 80.95$_{\pm2.33}$ & 88.71$_{\pm1.57}$ & 51.79$_{\pm3.04}$ & 53.35$_{\pm3.09}$ & 56.17$_{\pm2.95}$ & 56.48$_{\pm3.00}$ & 88.01$_{\pm1.97}$ & 93.27$_{\pm1.23}$ & 77.37$_{\pm2.50}$ & 88.31$_{\pm1.61}$  \\
Llama-70B\_2-shot-cot & 59.21$_{\pm2.86}$ & 60.06$_{\pm2.77}$ & 88.28$_{\pm1.97}$ & 94.33$_{\pm1.13}$ & 58.94$_{\pm3.04}$ & 60.20$_{\pm2.95}$ & 66.73$_{\pm2.77}$ & 66.86$_{\pm2.86}$ & 89.53$_{\pm1.79}$ & 93.97$_{\pm1.26}$ & 79.96$_{\pm2.50}$ & 89.38$_{\pm1.49}$  \\ 
Llama-70B\_3-shot-cot & 57.51$_{\pm2.86}$ & 58.47$_{\pm2.91}$ & 87.57$_{\pm1.88}$ & 93.66$_{\pm1.16}$ & 59.12$_{\pm3.04}$ & 60.35$_{\pm3.03}$ & 66.01$_{\pm2.77}$ & 66.50$_{\pm2.77}$ & 89.62$_{\pm1.88}$ & 94.07$_{\pm1.20}$ & 80.59$_{\pm2.50}$ & 89.79$_{\pm1.51}$ \\

     \bottomrule
\end{tabular}
    }
\caption{
EM and F1 scores of the models on our dataset, together with $95\%$-confidence intervals obtained from bootstrapping ($n = 1000$) on the dataset. It is noted that the scores from GPT-4 are based on 150 samples (similar to the subset used for human performance), while for others, they are based on the full version of MoreHopQA {\small w/ hv}. The baseline model is Llama-8B prompted with the full context and only the first two words of the question.
}
\label{tab_main_result}
\end{center}
\end{table*}

\subsection{Performance Categorization}
\label{subsec:performance categorization}

We present the details of the performance categorization in Table~\ref{tab:truth_table}.

\begin{table*}[h]
\centering
\begin{tabular}{ccccccl}

\toprule
\textbf{Case 1} & \textbf{Case 2} & \textbf{Case 3} & \textbf{Case 4} &\textbf{ Case 5 }&\textbf{ Case 6 }& \multicolumn{1}{l}{\textbf{Category}} \\
\midrule
\multirow{6}{*}{ T } & \multirow{5}{*}{T} & T & T & T & T & Perfect Reasoning \\
\cmidrule{3-7}
 &  &  \multicolumn{2}{c}{\multirow{3}{*}{ - }} & F & F & \multirow{3}{*}{Shortcut Reasoning } \\
& & \multicolumn{2}{c}{} & T & F & \\
& & \multicolumn{2}{c}{} & F & T & \\
\cmidrule{3-7}
& & \multicolumn{2}{c}{\textit{Either is F}} & T& T & Problematic Performance \\
\cmidrule{2-7}
& F &\multicolumn{4}{c}{-} & Problematic Performance \\
\midrule
\multirow{15}{*}{ F } & \multirow{7}{*}{ T } & \multicolumn{2}{c}{\multirow{3}{*}{ - }} & F & F & \multirow{3}{*}{Shortcut Reasoning } \\
& & \multicolumn{2}{c}{} & T & F & \\
& & \multicolumn{2}{c}{} & F & T & \\
\cmidrule{3-7}
& & T & F & T & T & Problematic Performance \\
\cmidrule{3-7}
& & - & T & T & T & Failed Reasoning \\
\cmidrule{3-7}
& & F & F & T & T & Extra Step Failure \\
\cmidrule{2-7}
& \multirow{9}{*}{ F } & \multirow{5}{*}{T} & F & F & \multirow{3}{*}{-} &  \multirow{3}{*}{Problematic Performance} \\
& & & T & F & & \\
& & & F & T & & \\
\cmidrule{4-7}
& & & \multirow{2}{*}{T} & \multirow{2}{*}{T} & T & Failed Reasoning \\
\cmidrule{6-7}
& & & &  & F &   Failure\\
\cmidrule{3-7}
\cmidrule{3-7}
& & \multirow{4}{*}{F} & T & T & - & Failed Reasoning \\
\cmidrule{4-7}
& & & F & F & \multirow{3}{*}{-} & \multirow{3}{*}{Failure} \\
& & & T & F &  & \\
& & & F & T &  & \\
\bottomrule
\end{tabular}
\caption{Categorizing the performance of the LLMs across various cases. T (true) means the LLM gives a correct answer to corresponding cases, while F (false) means the LLM gives a wrong one.}
\label{tab:truth_table}
\end{table*}

\end{document}